\documentclass[10pt,conference]{IEEEtran}
\IEEEoverridecommandlockouts
\usepackage{cite}
\usepackage{amsmath,amssymb,amsfonts}
\usepackage{algorithmic}
\usepackage{graphicx}
\usepackage{textcomp}
\usepackage{xcolor}
\def\BibTeX{{\rm B\kern-.05em{\sc i\kern-.025em b}\kern-.08em
    T\kern-.1667em\lower.7ex\hbox{E}\kern-.125emX}}
\usepackage{algorithm}
\usepackage{algorithmic}
\usepackage{booktabs}
\begin{document}

\title{MultiPress: A Multi-Agent Framework for \\Interpretable Multimodal News Classification}
\author{\textbf{Tailong Luo}$^{1}$,\textbf{Hao Li}$^{\IEEEauthorrefmark{2},2}$, \textbf{Rong Fu}$^{3}$,
\textbf{Xinyue Jiang}$^{4}$,
\textbf{Huaxuan Ding}$^{4}$,\\
\textbf{Yiduo Zhang}$^{4}$,
\textbf{Zilin Zhao}$^{4}$,
\textbf{Simon Fong}$^{3}$,
\textbf{Guangyin Jin}$^{5}$,
\textbf{Jianyuan Ni}$^{6}$\\
\textsuperscript{1}New York Institute of Technology 
\textsuperscript{2}University of Arizona 
 \textsuperscript{3}University of Macau \\
\textsuperscript{4}Peking University
\textsuperscript{5}Chang'an University
\textsuperscript{6}Juniata College
\thanks{\IEEEauthorrefmark{2} Corresponding author.}
}

\maketitle

\begin{abstract}

With the growing prevalence of multimodal news content, effective news topic classification demands models capable of jointly understanding and reasoning over heterogeneous data such as text and images. Existing methods often process modalities independently or employ simplistic fusion strategies, limiting their ability to capture complex cross-modal interactions and leverage external knowledge. To overcome these limitations, we propose \textit{MultiPress}, a novel three-stage multi-agent framework for multimodal news classification. \textit{MultiPress} integrates specialized agents for multimodal perception, retrieval-augmented reasoning, and gated fusion scoring, followed by a reward-driven iterative optimization mechanism. We validate \textit{MultiPress} on a newly constructed large-scale multimodal news dataset, demonstrating significant improvements over strong baselines and highlighting the effectiveness of modular multi-agent collaboration and retrieval-augmented reasoning in enhancing classification accuracy and interpretability.
\end{abstract}


\section{Introduction}

The rapid expansion of multimedia news content on digital platforms has transformed how information is produced and consumed. Modern news articles frequently combine textual narratives with images, videos, and other modalities to deliver richer and more engaging stories. This multimodal nature introduces new challenges for automated news topic classification, a critical task for news aggregation, recommendation systems, and public opinion analysis.

Traditional text-based classification methods fall short in fully exploiting the complementary information provided by visual content. Meanwhile, straightforward multimodal fusion techniques often struggle with modality imbalance, noisy inputs, and limited integration of external knowledge. Recent advances in Multimodal Large Language Models (MLLMs)~\cite{xi2023agentsurvey,li2023camel,wu2023autogen,madaan2023selfrefine,kang2025hssbench,feng2025seeing,liu2026reasonact,gao2025laobench} have shown strong potential for understanding heterogeneous data. However, directly applying MLLMs to multimodal news classification remains challenging due to difficulties in extracting and fusing cross-modal features, knowledge latency, hallucination, and the need for complex reasoning that jointly integrates multimodal inputs with external knowledge sources~\cite{ji2023survey,brown2020fewshot,kang2026multimodal}.

\begin{figure}
    \centering
    \includegraphics[width=1\linewidth]{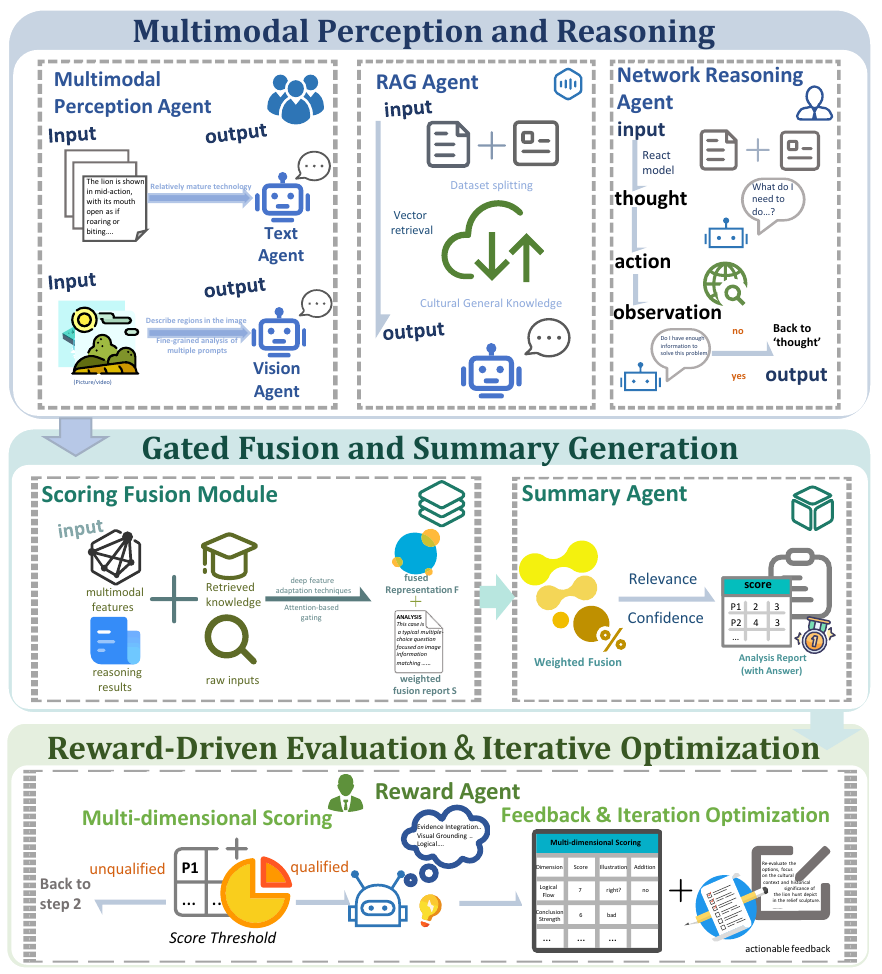}
    \vspace{-0.69cm}
    \caption{Overview of our MultiPress Framework.}
    \label{fig:framework}
    \vspace{-0.69cm}
\end{figure}

To address these challenges, we propose \textit{MultiPress}, a modular multi-agent framework for interpretable multimodal news classification. As illustrated in Figure~\ref{fig:framework}, specialized agents first analyze textual and visual inputs to extract modality-specific features. A Retrieval-Augmented Generation (RAG)~\cite{lewis2020rag,gao2023rag,kang2026quanteval} agent then enriches the context by retrieving relevant external knowledge via vector search~\cite{hong2023metagpt,bran2023chemcrow,he2026order,zhu2026edis}. Subsequently, an online reasoning agent iteratively refines the understanding by dynamically invoking MLLMs and web search, enabling more accurate and context-aware classification. The resulting signals are integrated through a gated fusion mechanism that adaptively weighs multimodal features and retrieved knowledge, while a reward-driven evaluation agent provides feedback for iterative refinement.

To rigorously evaluate our approach, we construct \textit{NewsMM}, a large-scale multimodal news dataset comprising over 7,000 articles paired with images and annotated into eight high-level topic categories. This dataset reflects real-world multimodal news characteristics and serves as a robust benchmark.

Extensive experiments demonstrate that \textit{MultiPress} significantly outperforms strong multimodal baselines. In addition to improving classification accuracy, \textit{MultiPress} explicitly provides interpretable evidence---including textual cues, visual concepts, and retrieved knowledge---which is important for debugging, auditability, and user trust in real-world news applications.

We will make our dataset and code publicly available upon acceptance. Our main contributions are summarized as follows:
\begin{itemize}
    \item We introduce \textit{MultiPress}, a novel multi-agent framework integrating multimodal perception, retrieval-augmented reasoning, adaptive fusion, and reward-driven optimization for robust multimodal news classification.
    \item We construct \textit{NewsMM}, a new large-scale multimodal news dataset with diverse topic categories.
    \item We conduct comprehensive experiments demonstrating that \textit{MultiPress} significantly outperforms existing baselines, validating the effectiveness of modular multi-agent collaboration.
\end{itemize}
\section{Method}

This section introduces \textit{MultiPress}, a three-stage multi-agent framework for interpretable multimodal news classification. 
Different from conventional multimodal classifiers that perform a single-pass fusion of text and images, MultiPress decomposes the task into collaborative agents specializing in perception, retrieval, reasoning, fusion, and reward-based refinement. 
Such modular decomposition enables stronger cross-modal grounding, reduces hallucination, and provides human-interpretable evidence for topic prediction. 
The overall workflow is illustrated in Figure~\ref{fig:framework}.

\subsection{Problem Formulation}

Given a multimodal news instance consisting of textual content $\mathbf{T}$ (headline and body) and an associated image $\mathbf{I}$, the goal is to predict its topic label
$y \in \{1,\dots,C\}$, where $C=8$ in \textit{NewsMM}. 
In addition to classification accuracy, MultiPress explicitly produces an interpretable fusion report $\mathbf{S}$ grounded in multimodal cues and retrieved evidence:

\begin{equation}
(\hat{y}, \mathbf{S}) = \mathcal{M}(\mathbf{T}, \mathbf{I}),
\end{equation}

where $\hat{y}$ is the predicted topic category and $\mathbf{S}$ contains supporting evidence, reasoning traces, and modality contributions.

\subsection{Overview of the Multi-Agent Design}

MultiPress consists of five specialized agents:

\begin{itemize}
    \item \textbf{Perception Agent} extracts modality-specific semantic cues.
    \item \textbf{RAG Agent} retrieves external knowledge evidence.
    \item \textbf{Online Reasoning Agent} performs iterative ReAct-style reasoning with search.
    \item \textbf{Fusion Agent} integrates multimodal and retrieved evidence via gated scoring.
    \item \textbf{Reward Agent} evaluates report quality and triggers refinement.
\end{itemize}

Each agent communicates through structured messages rather than free-form generations. 
This design improves controllability and interpretability.

\paragraph{Agent Communication Protocol.}
At each step, an agent receives a structured context:
\begin{equation}
\mathcal{C} = \{\mathbf{T}, \mathbf{I}, \mathbf{K}, \mathbf{R}, \mathbf{F}_{back}\},
\end{equation}

and outputs a structured message
\begin{equation}
m_a = \{\text{prediction},\text{evidence},\text{rationale},\text{confidence}\}.
\end{equation}

These messages are aggregated and passed across agents, enabling explicit modular collaboration.

\subsection{Stage 1: Multimodal Perception and Knowledge-Augmented Reasoning}

Stage 1 focuses on extracting modality-specific representations and enriching the context through retrieval and iterative reasoning.

\subsubsection{Multimodal Perception Agent}

The perception agent independently analyzes text and image inputs:
\begin{equation}
\mathbf{T}_{feat} = \mathcal{F}_T(\mathbf{T}), 
\quad 
\mathbf{I}_{feat} = \mathcal{F}_I(\mathbf{I}),
\end{equation}

where $\mathcal{F}_T$ captures entities, events, sentiment polarity, and discourse cues, while $\mathcal{F}_I$ extracts objects, scenes, and visual attributes.

To enhance interpretability, the agent outputs structured cues:
\begin{align}
\mathbf{Z}_T &= \{\text{entities},\text{keywords},\text{text-summary}\}, \\
\mathbf{Z}_I &= \{\text{objects},\text{scene},\text{image-summary}\}.
\end{align}

These intermediate representations provide explicit modality evidence for downstream fusion.

\subsubsection{Retrieval-Augmented Generation (RAG) Agent}

News classification often requires background knowledge that is not explicitly stated in the article, such as political figures, emerging events, or domain-specific terminology. 
Thus, the RAG agent retrieves evidence snippets from an external knowledge base:
\begin{equation}
\mathbf{K} = \mathcal{R}(\mathbf{T},\mathbf{I}).
\end{equation}

\paragraph{Cross-Modal Retrieval Scoring.}
We embed both modalities into a shared latent space. 
Let $\mathbf{e}_T$ and $\mathbf{e}_I$ denote embeddings of $\mathbf{T}$ and $\mathbf{I}$, and $\mathbf{e}_j$ denote the embedding of a candidate knowledge item $k_j$. 
A retrieval score is computed as:
\begin{equation}
s_j = \beta \cos(\mathbf{e}_T,\mathbf{e}_j) 
+ (1-\beta)\cos(\mathbf{e}_I,\mathbf{e}_j),
\end{equation}

where $\beta$ balances textual and visual relevance. The top-$k$ evidence items are returned:
\begin{equation}
\mathbf{K} = \{k_1,\dots,k_k\}.
\end{equation}

\paragraph{Knowledge Filtering.}
To avoid noisy retrieval, we discard snippets with similarity below a threshold and prioritize recent evidence for time-sensitive news topics.

\subsubsection{Online Reasoning Agent}

After retrieval, the online reasoning agent performs iterative reasoning inspired by ReAct. 
At iteration $t$, the agent generates a thought, executes an action (search or verification), and updates observations:
\begin{align}
\text{thought}_t &= \phi(\mathbf{T},\mathbf{I},\mathbf{K},\text{obs}_{t-1}), \\
\text{action}_t &= \psi(\text{thought}_t), \\
\text{obs}_t &= \omega(\text{action}_t).
\end{align}

The reasoning trace is summarized as:
\begin{equation}
\mathbf{R} = \{\text{step-wise rationale with citations}\}.
\end{equation}

\paragraph{Stopping Criteria and Budget Control.}
To ensure efficiency, reasoning terminates early if:
(i) the predicted label distribution stabilizes across consecutive iterations, or
(ii) the reward score exceeds threshold $\tau$.
We limit the loop to at most $N$ iterations and $M$ search actions per iteration.

\subsection{Stage 2: Gated Fusion and Report Generation}

Stage 2 integrates multimodal features and retrieved evidence into a unified representation.

\subsubsection{Adaptive Gated Fusion}

We compute modality-aware gating weights:
\begin{equation}
[\alpha_T,\alpha_I,\alpha_K,\alpha_R] =
\mathrm{softmax}(W_g[\mathbf{T}_{feat};\mathbf{I}_{feat};\mathbf{K};\mathbf{R}]).
\end{equation}

The fused representation is:
\begin{equation}
\mathbf{F}=\alpha_T\mathbf{T}_{feat}+\alpha_I\mathbf{I}_{feat}
+\alpha_K\mathbf{K}+\alpha_R\mathbf{R}.
\end{equation}

\paragraph{Reliability-Aware Fusion.}
The gating mechanism implicitly reflects evidence reliability. For example, if the image is a generic stock photo or retrieval confidence is low, the model assigns smaller $\alpha_I$ or $\alpha_K$, thereby mitigating modality noise.

\subsubsection{Fusion Report Generation}

A summary assistant generates the final interpretable report:
\begin{equation}
\mathbf{S} = \mathcal{H}(\mathbf{F}),
\end{equation}

where $\mathbf{S}$ includes: predicted topic, supporting multimodal cues, retrieved evidence, and explanation.

\subsection{Stage 3: Reward-Driven Evaluation and Iterative Refinement}

Stage 3 introduces a reward agent that evaluates report quality and triggers refinement.

\subsubsection{Reward Function}

The reward is computed as:
\begin{equation}
r = \lambda_1 r_{cls} + \lambda_2 r_{ground} + \lambda_3 r_{cons}.
\end{equation}

Here, $r_{cls}$ measures classification confidence, $r_{ground}$ evaluates evidence grounding, and $r_{cons}$ measures cross-modal consistency. We set $\lambda_1=0.5,\lambda_2=0.3,\lambda_3=0.2$.

\paragraph{Reward Components.}
$r_{cls}$ is computed as the softmax margin between the top-1 and top-2 predicted topics. 
$r_{ground}$ measures the proportion of rationale sentences that can be semantically matched to retrieved evidence. 
$r_{cons}$ evaluates agreement between key entities extracted from text and objects extracted from images.

\paragraph{Iterative Refinement.}
If $r<\tau$, the reward agent generates feedback:
\begin{equation}
\mathbf{F}_{back} = \mathcal{F}_{back}(r,\mathbf{S}),
\end{equation}

which is sent back to the fusion stage for refinement until convergence.

\subsection{Complexity Analysis}

MultiPress introduces additional overhead due to retrieval and iterative reasoning. 
Let $k$ be the retrieval depth and $N$ the maximum reasoning iterations. The overall inference cost scales as:
\begin{equation}
\mathcal{O}(k + N\cdot M),
\end{equation}

which remains practical for offline news classification and batch processing scenarios.

\begin{algorithm}[!t]
\caption{MultiPress Multi-Agent Inference Pipeline}
\label{alg:multipress}
\small
\begin{algorithmic}[1]
\REQUIRE News text $\mathbf{T}$, image $\mathbf{I}$
\STATE Extract features $\mathbf{T}_{feat},\mathbf{I}_{feat}$
\STATE Retrieve evidence $\mathbf{K}=\mathcal{R}(\mathbf{T},\mathbf{I})$
\FOR{$t=1$ to $N$}
    \STATE Online reasoning with search feedback $\mathbf{R}_t$
    \STATE Fuse evidence via gated fusion $\mathbf{F}_t$
    \STATE Generate report $\mathbf{S}_t=\mathcal{H}(\mathbf{F}_t)$
    \STATE Compute reward score $r_t=\mathcal{E}(\mathbf{S}_t)$
    \IF{$r_t>\tau$}
        \STATE \textbf{break}
    \ENDIF
\ENDFOR
\STATE Output final prediction $\hat{y}$ and report $\mathbf{S}_t$
\end{algorithmic}
\end{algorithm}

\subsection{Summary}
Algorithm~\ref{alg:multipress} summarizes the full MultiPress inference procedure.
Overall, MultiPress decomposes multimodal news classification into modular agent collaboration:

\begin{equation}
(\hat{y},\mathbf{S})=
\mathcal{H}\Big(\mathcal{G}(\mathcal{F}_T,\mathcal{F}_I,\mathcal{R},\mathcal{O})\Big),
\end{equation}

enabling robust multimodal understanding, external knowledge grounding, and interpretable reasoning through reward-driven refinement.

\section{The NewsMM Dataset}

\textbf{Data Collection.} We construct \textit{NewsMM} by crawling multimodal news articles from reputable online news portals and social media platforms. Each instance includes a news article’s headline and body text paired with one or more relevant images. To ensure diversity, the dataset covers eight topic categories: \textit{Politics, Economy, Technology, Sports, Entertainment, Health, Environment,} and \textit{Science}.

Textual content is preprocessed to remove advertisements and unrelated metadata. Images are filtered to retain only those directly relevant to the news content. Articles lacking quality images are excluded to maintain dataset integrity.

\textbf{Annotation Process.} A team of 10 annotators with journalism and media expertise labels each sample with one of the eight topic categories based on combined textual and visual content. Each sample is independently labeled by three annotators; majority voting determines the final label. Disagreements are resolved through expert review.

To enhance consistency and reduce bias, a GPT-based topic prediction model is used as a complementary annotator. Samples with conflicting human and GPT labels undergo further manual verification or are discarded if ambiguity persists.

\textbf{Data Statistics.}
\textit{NewsMM} contains 7,200 multimodal news samples evenly distributed across eight categories (900 per category). The dataset is split into 90\% training (6,480 samples) and 10\% testing (720 samples).

Textual data averages 45.2 tokens per headline and 320.7 tokens per article body. Images are standardized to 256×256 pixels, with an average of 1.8 images per article.

\textbf{Knowledge Base Construction.}
To support retrieval-augmented reasoning, we build a comprehensive external knowledge base (KB) tailored for news classification, integrating:

\begin{itemize}
    \item \textbf{Multimodal News Samples:} All \textit{NewsMM} training samples with text, images, and labels.
    \item \textbf{External News Corpora:} Large-scale textual datasets such as AG News, Reuters-21578, and CNN.
    \item \textbf{Domain-Specific Knowledge:} Curated knowledge graphs and encyclopedic entries related to current events, organizations, and entities.
\end{itemize}

The KB is indexed using a vector database optimized for fast semantic retrieval. Textual and visual features are embedded into a shared latent space via pretrained multimodal encoders, enabling efficient cross-modal similarity search. This design allows the RAG agent to retrieve highly relevant knowledge snippets that enrich classification context.
\paragraph{Category definition.}
We define eight high-level topics following common news taxonomies, where each label covers a broad yet coherent semantic scope.
For example, \textit{Politics} includes elections, diplomacy, and government policies; \textit{Economy} covers markets, finance, trade, and macroeconomic indicators; \textit{Technology} focuses on consumer tech, AI, and industrial innovation; and \textit{Environment} includes climate events, conservation, and energy transitions.
During annotation, we provide annotators with concise label guidelines and borderline examples to reduce ambiguity between closely related categories (e.g., \textit{Science} vs. \textit{Technology}, \textit{Health} vs. \textit{Science}).

\paragraph{Quality control and agreement.}
To ensure annotation reliability, each sample is labeled by three annotators, followed by majority voting.
We additionally conduct spot checks on randomly sampled subsets and perform guideline refinement when systematic confusion is observed.
As an agreement signal, we compute inter-annotator consistency based on the proportion of samples with unanimous votes and those requiring adjudication.
In practice, ambiguous instances often arise when images are generic (e.g., portraits or stock photos) or when the article narrative spans multiple domains; such cases are either resolved by expert review or removed if the topic remains uncertain.



\section{Experiments}

\subsection{Experimental Settings}

We evaluate \textit{MultiPress} on \textit{NewsMM} using GPT-4o~\cite{openai2024gpt4ocard} and Qwen2.5-VL-7B-Instruct~\cite{bai2025qwen25vltechnicalreport} as backbone multimodal models. 
We report overall accuracy (Acc.), macro-averaged precision (MP), recall (MR), and F1-score (MF1) to assess performance across all categories.

Unless otherwise specified, the multi-agent pipeline follows a unified configuration: 
(i) the perception agent extracts text/image cues, 
(ii) the retrieval agent queries a FAISS-based vector index with top-$k=5$, 
(iii) the online reasoning agent performs up to 3 ReAct-style iterations with 2--3 search steps per iteration, and 
(iv) the reward agent evaluates outputs along accuracy, groundedness, and cross-modal consistency, with an acceptance threshold $\tau=0.85$. 
All methods share the same train/test split for fair comparison.

\subsection{Baseline Methods}

We compare \textit{MultiPress} with a broad set of strong multimodal baselines covering both open-source and proprietary models.
Our primary baselines include GLM-4V-9B~\cite{glm2024chatglmfamilylargelanguage}, mPLUG-7B~\cite{ye2024mplugowlmodularizationempowerslarge}, Qwen2.5-VL-3B/7B-Instruct (zero-shot and fine-tuned), and GPT-4o.


\subsection{Overall Performance on NewsMM}

\begin{table}[!t]
\centering
\small
\vspace{-0.2cm}
\caption{Performance comparison on \textit{NewsMM} test set.}
\vspace{-0.2cm}
\label{tab:overall_performance_news}
\setlength{\tabcolsep}{4pt}
\renewcommand{\arraystretch}{0.8}
\begin{tabular}{lcccc}
\toprule
\textbf{Model} & \textbf{Acc.} & \textbf{MP} & \textbf{MR} & \textbf{MF1} \\
\midrule
GLM-4V-9B & 70.5 & 69.8 & 69.2 & 69.5 \\
mPLUG-7B & 68.9 & 68.1 & 67.5 & 67.8 \\
Qwen2.5-VL-3B (zero-shot) & 62.3 & 61.0 & 60.5 & 60.7 \\
Qwen2.5-VL-3B (fine-tuned) & 66.8 & 66.0 & 65.5 & 65.7 \\
Qwen2.5-VL-7B (zero-shot) & 77.4 & 76.8 & 76.1 & 76.4 \\
Qwen2.5-VL-7B (fine-tuned) & 80.2 & 79.5 & 79.0 & 79.2 \\
GPT-4o-20250513 & 79.6 & 79.0 & 78.5 & 78.7 \\
\midrule
\textbf{MultiPress (GPT-4o)} & \textbf{91.2} & \textbf{90.7} & \textbf{90.1} & \textbf{90.4} \\
\textbf{MultiPress (Qwen2.5-VL-7B)} & 85.3 & 84.7 & 84.1 & 84.4 \\
\bottomrule
\end{tabular}
\vspace{-0.2cm}
\end{table}

Table~\ref{tab:overall_performance_news} summarizes the main results on \textit{NewsMM}. 
\textit{MultiPress} consistently outperforms strong multimodal baselines. 
Notably, fine-tuning Qwen2.5-VL-7B improves zero-shot accuracy from 77.4\% to 80.2\%, but still trails \textit{MultiPress (Qwen2.5-VL-7B)} by 5.1 points and \textit{MultiPress (GPT-4o)} by 11.0 points. 
This indicates that the improvement is not merely from stronger backbones or fine-tuning, but from the \emph{multi-agent decomposition} that explicitly enables knowledge grounding and iterative correction.

\subsection{Generalization Beyond News Domain}

To validate whether MultiPress generalizes beyond news classification, we also apply it on three multimodal reasoning benchmarks using Qwen2.5-VL-7B-Instruct as the backbone. 
MultiPress improves accuracy from 24.05 to 31.42 on MathVerse, from 54.00 to 61.35 on MMStar, and from 19.80 to 26.17 on MMMU. 
These consistent gains (roughly +6--7 points) suggest that the proposed agentic retrieval and reasoning mechanism provides general benefits for multimodal understanding, rather than overfitting to the news domain.

\subsection{Ablation Study}

\begin{table}[!t]
\centering
\vspace{-0.2cm}
\caption{Ablation study on \textit{NewsMM} test set.}
\setlength{\tabcolsep}{4pt}
\renewcommand{\arraystretch}{0.8}
\vspace{-0.2cm}
\label{tab:ablation_news}
\begin{tabular}{lcc}
\toprule
\textbf{Variant} & \textbf{Acc.} & \textbf{MF1} \\
\midrule
Full \textit{MultiPress} (GPT-4o) & 91.2 & 90.4 \\
\quad w/o Retrieval-Augmented Generation & 86.5 & 85.7 \\
\quad w/o Online Reasoning Agent & 87.1 & 86.3 \\
\quad w/o Gated Fusion & 84.9 & 84.1 \\
\quad w/o Reward-Driven Optimization & 88.3 & 87.5 \\
\quad Text-only input & 79.4 & 78.6 \\
\quad Image-only input & 72.8 & 71.9 \\
\bottomrule
\end{tabular}
\vspace{-0.2cm}
\end{table}

Table~\ref{tab:ablation_news} shows that each module contributes meaningfully to the final performance. 
Removing the RAG agent drops accuracy by 4.7 points, verifying that external knowledge grounding is crucial for long-tail entities and emerging events. 
Disabling the online reasoning agent causes a 4.1-point decrease, indicating that iterative search-and-reflection can correct initial misunderstandings or incomplete evidence. 
The largest degradation comes from removing gated fusion, showing that adaptive modality weighting is essential for suppressing noisy images and resolving modality conflicts. 
Finally, the reward-driven optimization contributes +2.9 points, suggesting that explicit multi-criteria evaluation helps refine both the prediction and the interpretability report.

\subsection{Effect of Reward-Driven Iteration}

We further examine how many refinement iterations are necessary. 
With 0 iterations (single pass), MultiPress achieves 87.5\% Acc. and 86.8\% MF1. 
Allowing 1 iteration improves performance to 90.1\% Acc. / 89.3\% MF1, and 2 iterations reaches the best result of 91.2\% Acc. / 90.4\% MF1. 
Using 3 iterations yields marginal change, indicating that most gains are obtained within the first 1--2 refinement rounds and additional iterations show diminishing returns. 
This observation supports our design choice of using a small iteration budget for practical deployment.

\subsection{Retrieval Sensitivity Analysis}

We study how retrieval depth influences performance. 
When retrieval is shallow (top-$k=1$), MultiPress obtains 89.4\% Acc. / 88.6\% MF1. 
Increasing to top-$k=3$ improves to 90.8\% Acc. / 89.9\% MF1, and top-$k=5$ achieves the best result of 91.2\% Acc. / 90.4\% MF1. 
However, using top-$k=10$ slightly degrades performance (90.5\% Acc. / 89.7\% MF1), suggesting that overly large retrieval depth may introduce irrelevant or conflicting evidence, increasing reasoning burden. 
Overall, MultiPress remains robust across a reasonable range of $k$, and benefits most from moderate retrieval depth.

\subsection{Interpretability Evaluation}

\begin{table}[!t]
\centering
\small
\vspace{-0.15cm}
\caption{Interpretability evaluation on 100 sampled test instances (higher is better).}
\vspace{-0.15cm}
\label{tab:interp}
\setlength{\tabcolsep}{2pt}
\renewcommand{\arraystretch}{0.85}
\begin{tabular}{lccc}
\toprule
\textbf{Model} & \textbf{Faithfulness} & \textbf{Groundedness} & \textbf{Usefulness} \\
\midrule
Qwen2.5-VL-7B & 3.42 & 3.15 & 3.38 \\
GPT-4o-20250513 & 4.10 & 3.85 & 4.45 \\
\textbf{MultiPress (GPT-4o)} & \textbf{4.65} & \textbf{4.72} & \textbf{4.58} \\
\bottomrule
\end{tabular}
\vspace{-0.25cm}
\end{table}

Beyond accuracy, MultiPress explicitly targets interpretability. 
We randomly sample 100 test instances and ask three annotators to score model-generated rationales on a 1--5 Likert scale across faithfulness (to inputs), groundedness (to retrieved evidence), and usefulness (for human understanding). 
Table~\ref{tab:interp} shows that MultiPress yields substantially higher groundedness (4.72 vs. 3.85 for direct GPT-4o), confirming that retrieval and online reasoning reduce hallucination by encouraging evidence-backed explanations. 
Meanwhile, the improvement in faithfulness indicates that gated fusion helps align rationales with modality-specific cues rather than producing generic explanations.

\subsection{Efficiency Analysis}

\begin{table}[!t]
\centering
\small
\vspace{-0.15cm}
\caption{Efficiency comparison averaged per sample.}
\vspace{-0.15cm}
\label{tab:eff}
\setlength{\tabcolsep}{2pt}
\renewcommand{\arraystretch}{0.85}
\begin{tabular}{lccc}
\toprule
\textbf{Method} & \textbf{\#Calls} & \textbf{Tokens} & \textbf{Latency} \\
\midrule
GPT-4o (direct) & 1.0 & 1,450 & 1.8s \\
MultiPress (GPT-4o) & 6.2 & 5,820 & 8.5s \\
\bottomrule
\end{tabular}
\vspace{-0.25cm}
\end{table}

MultiPress introduces additional overhead due to retrieval and iterative reasoning. 
As shown in Table~\ref{tab:eff}, compared with direct prompting, MultiPress uses more model calls and tokens, increasing latency from 1.8s to 8.5s per sample. 
However, we note that the evaluation setting includes online search and multi-step refinement, which primarily targets improved robustness and interpretability rather than minimal latency. 
In practice, the framework is well-suited for offline/batch news processing, and future work can explore distillation or caching strategies to reduce inference cost.



\section{Related Work}

\subsection{Multimodal News Classification}

Multimodal news classification benefits from integrating textual and visual cues, showing improvements over text-only methods~\cite{wang-etal-2022-n24news}. While traditional multimodal methods use co-attention and joint embeddings, they struggle with incorporating external knowledge and performing complex reasoning.

\subsection{Retrieval-Augmented Multimodal Models}

Retrieval-Augmented Generation (RAG) enhances models by retrieving external knowledge, extending to multimodal settings for improved grounding and reducing hallucinations~\cite{lewis2020rag,mei2025surveymultimodalretrievalaugmentedgeneration}. Recent agentic RAG frameworks, such as HM-RAG and MultiRAG, better coordinate retrieval and reasoning across modalities~\cite{liu2025hm,mao2025multiragmultimodalretrievalaugmentedgeneration}.

\subsection{Multi-Agent Reasoning and Fusion}

Multi-agent systems have been applied to multimodal reasoning by decomposing tasks into specialized agents. These systems enable effective knowledge synthesis and dynamic fusion, which enhances cross-modal reasoning~\cite{liu2025hm}. Our work builds on this by integrating multimodal perception, retrieval, reasoning, fusion, and refinement into a unified framework for interpretable news classification.

\section{Conclusion}

We present \textit{MultiPress}, a novel three-stage multi-agent framework for multimodal news topic classification. By decomposing the task into specialized agents for multimodal perception, retrieval-augmented reasoning, gated fusion, and reward-driven optimization, \textit{MultiPress} effectively captures complex cross-modal interactions and integrates external knowledge to improve accuracy and robustness.
To support evaluation, we construct \textit{NewsMM}, a large-scale multimodal news dataset with over 7,000 annotated samples across eight topics. Extensive experiments demonstrate \textit{MultiPress} significantly outperforms strong baselines, validating the benefits of modular multi-agent collaboration and retrieval-augmented reasoning.

Future work will explore extending to additional modalities such as video and audio, incorporating real-time news streams, and investigating advanced reasoning strategies to further enhance multimodal news understanding.

\bibliographystyle{IEEEbib}
\bibliography{strings}

\end{document}